\documentclass[lettersize,journal]{IEEEtran}
\usepackage{amsmath,amsfonts}
\usepackage{algorithmic}
\usepackage{algorithm}
\usepackage{array}
\usepackage{multirow}
\usepackage{textcomp}
\usepackage{stfloats}
\usepackage{url}
\usepackage{verbatim}
\usepackage{graphicx}
\usepackage{cite}
\usepackage{xcolor}
\usepackage{colortbl, booktabs}
\usepackage{graphics} % for pdf, bitmapped graphics files
\usepackage{epsfig} % for postscript graphics files
\usepackage{mathptmx} % assumes new font selection scheme installed
\usepackage{times} % assumes new font selection scheme installed
\usepackage{amsmath} % assumes amsmath package installed
\usepackage{amssymb}  % assumes amsmath package installed
\usepackage{lipsum}
\usepackage{adjustbox}

\usepackage{booktabs} 
\usepackage{mathptmx}
\usepackage[mathcal]{euscript}
\usepackage{algorithm}
\usepackage{algorithmic}
\usepackage{amsmath}
\usepackage{amssymb}
\usepackage{soul,color,xcolor}      % 用于设置字体颜色
\definecolor{myColor}{rgb}{0,0,0}        % 隐藏修订时用这行
\makeatletter
\DeclareRobustCommand*{\revise}{\@ifnextchar\bgroup{\revise@}{\color{myColor}}}
\newcommand*{\revise@}[1]{{\textcolor{myColor}{#1}}}
\makeatother

\begin{document}

\title{\revise{Align then Adapt: Rethinking Parameter-Efficient Transfer Learning in 4D Perception}}

\author{Yiding Sun, Jihua Zhu, Haozhe Cheng, Chaoyi Lu, Zhichuan Yang, Lin Chen, and Yaonan Wang
        % <-this % stops a space
\thanks{This work was supported in part by NSFC (No. 62125305) , the Natural Science Basis Research Plan in Shaanxi Province of China (No. 2025JC-JCQN-091) and Technology Innovation Leading Program of Shaanxi (Program No. 2024QY-SZX-23).}% <-this % stops a space
\thanks{Yiding Sun, Jihua Zhu, Haozhe Cheng, Chaoyi Lu, Zhichuan Yang and Lin Chen are with the School of Software Engineering, Xi'an Jiaotong University, Xi'an 710048, China; \revise{Yiding Sun and Jihua Zhu are also with the State Key Laboratory of Human-Machine Hybrid Augmented Intelligence, Xi'an Jiaotong University, Xi'an 710048, China} (e-mail:  zhujh@xjtu.edu.cn) \textit{(Corresponding author: Jihua Zhu)}.}
\thanks{Yaonan Wang is with the School of Electrical and Information Engineering, Hunan University, Changsha, 410082, China, and also with the National Engineering Research Center for Robot Visual Perception and Control Technology, Changsha 410082, China (e-mail: yaonan@hnu.edu.cn).}
}

\markboth{Journal of \LaTeX\ Class Files,~Vol.~14, No.~8, August~2021}%
{Shell \MakeLowercase{\textit{et al.}}: A Sample Article Using IEEEtran.cls for IEEE Journals}


\maketitle

\begin{abstract}
Point cloud video understanding is critical for robotics as it accurately encodes motion and scene interaction. We recognize that 4D datasets are far scarcer than 3D ones, which hampers the scalability of self-supervised 4D models. A promising alternative is to transfer 3D pre-trained models to 4D perception tasks. However, rigorous empirical analysis reveals two critical limitations that impede transfer capability: overfitting and the modality gap. To overcome these challenges, we develop a novel ``Align then Adapt” (PointATA) paradigm that decomposes parameter-efficient transfer learning into two sequential stages. Optimal-transport theory is employed to quantify the distributional discrepancy between 3D and 4D datasets, enabling our proposed point align embedder to be trained in Stage 1 to alleviate the underlying modality gap. To mitigate overfitting, an efficient point-video adapter and a spatial-context encoder are integrated into the frozen 3D backbone to enhance temporal modeling capacity in Stage 2. Notably, with the above engineering-oriented designs, PointATA enables a pre-trained 3D model without temporal knowledge to reason about dynamic video content at a smaller parameter cost compared to previous work. Extensive experiments show that PointATA can match or even outperform strong full fine-tuning models, whilst enjoying the advantage of parameter efficiency, \textit{e.g.} 97.21\% accuracy on 3D action recognition, +8.7\% on 4D action segmentation, \revise{and 84.06\% on 4D semantic segmentation.} 

\end{abstract}


\section{Introduction}
\IEEEPARstart{P}{oint} cloud videos, merging 3D space with 1D time, are indispensable for achieving spatial intelligence. It is crucial to clearly understand 4D videos for sensing environmental changes and interacting with the real world. This stands in sharp contrast to the \revise{limited descriptive power of 2D images and static 3D point clouds~\cite{SUN2026112800}.} Consequently, 4D representation learning has attracted broad research interest in recent years, with applications spanning robotics, augmented and virtual reality (AR\&VR) and beyond.

Current 4D point cloud video representation learning methods fall into two main categories: end-to-end supervised learning~\cite{10377208,liu2025mamba4d} and self-supervised pre-training followed by fine-tuning~\cite{shen2023pointcmp,deng2024vg4d,Shen_2023_ICCV}. However, almost all existing methods focus on the scenario of one network with task-specific weights, where a single model is trained from scratch or fully fine-tuned for each dataset. As the number of tasks increases, maintaining a distinct set of weights for every dataset becomes impractical. Moreover, 4D point cloud videos are two orders of magnitude smaller than their static 3D counterparts~\cite{chang2015shapenet,uy2019revisiting,wu20153d} and are more difficult to collect and manage. Under these constraints, reusing well-established 3D pre-trained models to guide cross-modal 4D perception could invigorate community research and advance the goal of universal representation.

\begin{figure}[t]
\centering
\includegraphics[width=1.0\columnwidth]{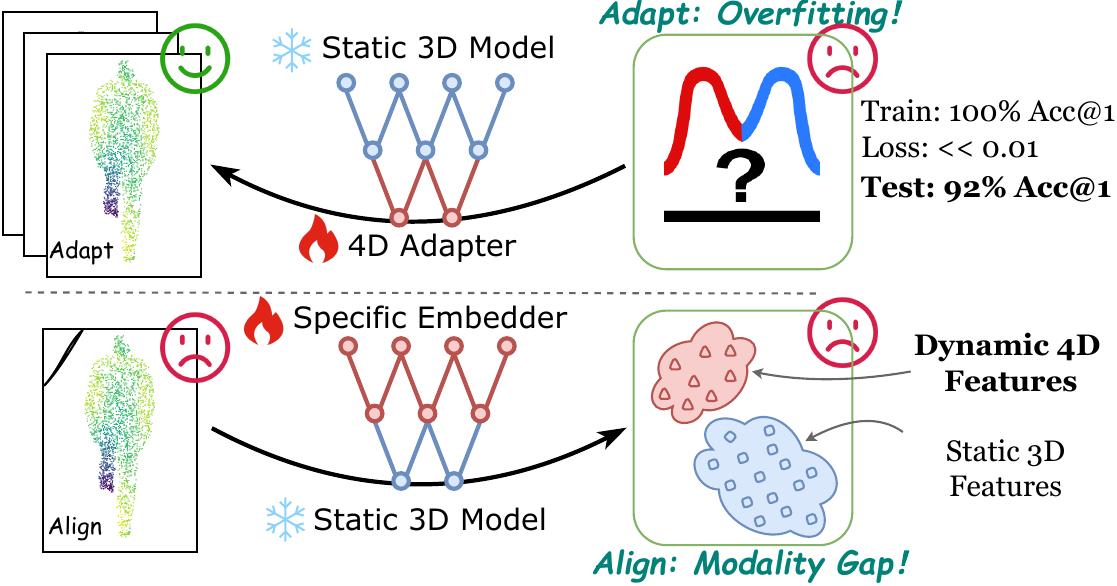}
\caption{\textbf{Current 4D PETL methods face two limits.}\textit{ Upper panel}: Adapters are feasible (Smiley), yet current methods expose models to severe overfitting (Crying). \textit{Lower panel}: Cross-modal transfer needs prior alignment. This practice is mature in 2D Vision and NLP. But to our knowledge, no 4D PETL study measures this gap (Crying). Without considering the gap, it will hurt downstream performance undoubtedly (Crying).}
\label{ill}
\end{figure}

In this paper, we investigate a novel, critical problem: \textbf{How can we surmount transfer barrier to efficiently adapt 3D foundation models to data-scarce 4D tasks?} Considering that training 4D models are drastically more expensive in both computing resource and time than 3D models, tackling this problem is both timely and valuable in practice. As a pioneer,~\cite{11093957} introduced the ``additive 4D Adapter” paradigm. It designs an adaptation strategy that incorporates temporal awareness and achieves strong performance on several 4D tasks. However, is this paradigm truly optimal? Our in-depth analysis in Fig.~\ref{ill} reveals two critical limitations in current parameter-efficient transfer learning (PETL) approaches for 4D perception that greatly hinder further progress. 

\begin{figure*}[t]
\centering
\includegraphics[scale=0.76]{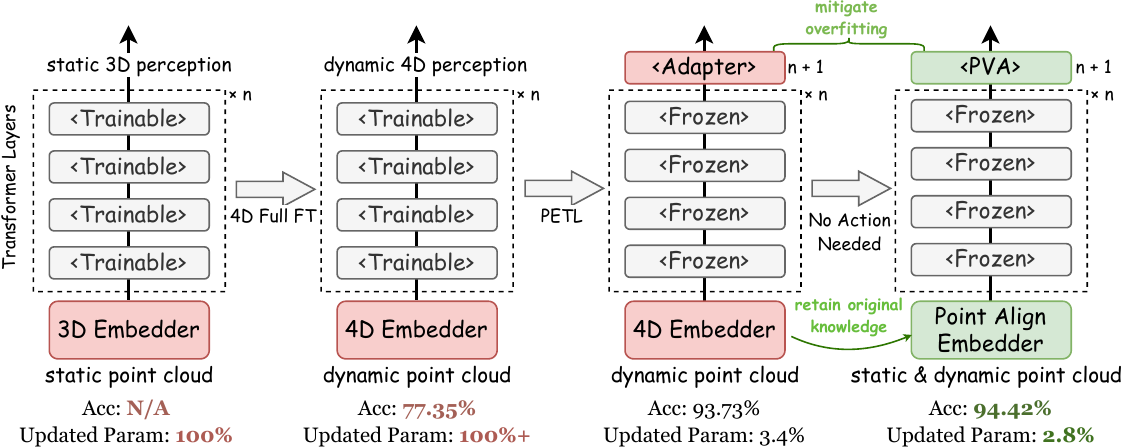}
\caption{\textbf{Comparison of 3D full fine-tuning, 4D full fine-tuning, 4D adapter tuning, and our PointATA.} \revise{Considering that the embedder required for 4D encoding is often heavier than that for 3D, when facing a 4D dynamic perception task, the number of parameters that need to be updated is even greater (over 100\%) than the full update for 3D.} PointATA saves large amount of resource and time than full fine-tuning and significantly boosts reuse of 3D pre-trained models. It also exploits 3D priors better than 4D adapter tuning and further cuts parameters to curb overfitting.}
\label{fig2}
\end{figure*}

First, \textbf{directly attaching a 4D adapter to a frozen 3D model for fine-tuning is intuitive yet naive.} During training, we observe that the network overfits heavily from the beginning of fine-tuning. We attribute this to the fact that pre-trained 3D models cannot infer temporal structure, which forces the adapters to fit spurious details and noise in 4D data. Second,~\cite{ma2024learning} in 2D Vision observe that fine-tuning Swin Transformer~\cite{liu2021Swin} on some target modalities helps the encoder extract more discriminative features while others hurts. This empirical finding indicates that\textbf{ distinct modality-specific semantic knowledge exists and affects cross-modal transfer to varying degrees.} It is widely accepted that static and dynamic point cloud, like image and video, form a modality pair and exhibit a modality gap~\cite{pan2022st,yangaim}. But to our knowledge, no prior work has explicitly measured this gap. When such knowledge conflicts, the representations learned by 3D pre-trained models remain under-utilized, and 4D PETL performance hurts inevitably.

We delve deeper into these two limitations and propose ``Align then Adapt” (PointATA), a paradigm that leverages 3D priors for 4D perception, as shown in Fig.~\ref{fig2}. Unlike conventional PETL, PointATA is a two-stage fine-tuning scheme that mitigates overfitting and fully releases the capacity of the pre-trained 3D model. We introduce two key components: (i) Enlightened by Optimal Transport (OT)~\cite{alvarez2020geometric}, we design a point align embedder. After resolving the dimensionality mismatch, we keep the 4D embedder learnable and use a frozen mini-PointNet~\cite{qi2017pointnet} as the 3D embedder during Stage 1. The objective is to minimize the distance between the joint distributions of 4D and 3D embeddings, ensuring that the pre-trained model performs effectively. (ii) In Stage 2, since the 3D model is already powerful, we introduce an engineering-oriented point video adapter (PVA) with spatial context encoder (SCE). SCE replicates the original multilayer perceptron (MLP), and the PVA uses a bottleneck structure with depth-wise separable convolution to limit parameters. 

We evaluate PointATA on multiple tasks, achieving \revise{+1.79\% recognition accuracy on MSR-Action3D~\cite{5543273}}, +8.7\% segmentation accuracy on HOI4D~\cite{Liu_2022_CVPR} and +0.9\% mIoU on Synthia 4D~\cite{choy20194d}. We also perform comprehensive ablation studies to demonstrate that PointATA could leverage knowledge from a large 3D model and achieves superior 4D perception with far fewer parameters.

We summarize our contributions as follows:
\begin{itemize}
    \item We reveal the limitation of PETL in 4D perception and introduce the PointATA paradigm, which preserves static-dynamic alignment and improves generalization.
    \item PointATA transfers priors from static 3D models to 4D domains via embedding alignment. We further propose engineering-oriented PVA with SCE to exploit large pre-trained 3D models more efficiently.
    \item Extensive experiments on various benchmarks demonstrate the superior efficiency and effectiveness of PointATA, outperforming existing methods in both accuracy and resource utilization.
\end{itemize}

\begin{figure*}[t]
\centering
\includegraphics[scale=0.9]{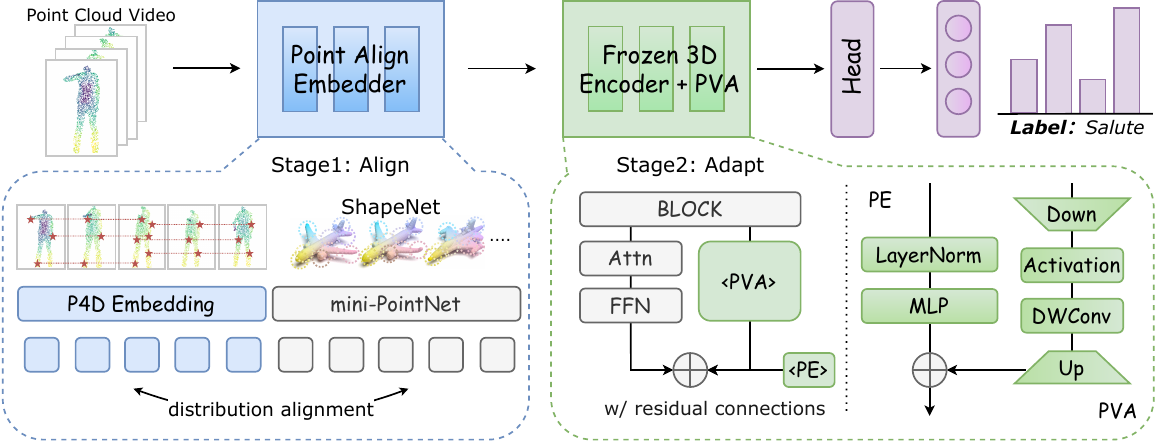}
\caption{\textbf{PointATA employs a two-stage workflow to quickly adapt large 3D pre-trained models to diverse 4D downstream tasks.} In Stage 1, it obtains 4D features via the P4D embedder and learns by minimizing the distribution distance to 3D source features. The weight of P4D embedder is randomly initialized. In Stage 2, it jointly fine-tunes the P4D embedder and the PVA to minimize task loss. The 3D backbone remains frozen throughout.}
\label{pipeline}
\end{figure*}

% \vspace{-0.25cm}
\section{RELATED WORK}
\subsection{Pre-trained Point Cloud Model}
To advance spatial perception, the community has developed a series of self-supervised 3D pre-trained models that boost downstream performance. Thanks to these ongoing efforts, 3D pre-trained models that were once scarce are now steadily scaling to new peaks~\cite{9913730,9891826,9855233,10249213}. We now have numerous powerful 3D pre-trained models, reflecting the continuous refinement of 3D pretext tasks~\cite{hu2024hyperbolic,cheng_edgcnet_2024}. Current pretext tasks fall into two groups: discriminative and generative. Discriminative tasks build positive and negative pairs to minimize the distance to the most similar neighbor in feature space. PointContrast~\cite{PointContrast2020} and CrossPoint~\cite{afham2022crosspoint} have advanced both in-modality and cross-modal learning respectively. Generative tasks learn latent features through autoregressive or autoencoding models. PointMAE~\cite{pang2022masked} learns powerful 3D representations by reconstructing heavily masked tokens. PointGPT~\cite{chen2024pointgpt} predicts the next token autoregressively, yielding scalable and data-efficient 3D representations. However, we believe that the utility of these pre-trained models extends far beyond 3D perception. Our work enhances their reusability and confirms a promising path from 3D pre-training to 4D perception.

\subsection{Point Cloud Video Perception}
Point cloud videos are unordered in space yet ordered in time. This duality requires 4D models to fuse temporal and spatial cues to capture both geometry and dynamics. Early methods rely on voxel grids. MinkowskiNet~\cite{choy20194d} voxelizes the points and applies 4D convolutions, but this incurs heavy information loss and high latency. Therefore, the community has shifted to point-based approaches. MeteorNet~\cite{9010250} first adds time dimension and tracks motion at the point level. P4Transformer~\cite{fan21p4transformer}, PPTr~\cite{wen2022point}, and LeaF~\cite{10377208} craft various transformers to enlarge receptive fields. Following the 3D success path, some recent efforts explore self-supervised 4D pre-training. C2P~\cite{zhang2023complete} proposes complete-to-part distillation over short clips. MaST-Pre~\cite{shen2023masked} embeds point cloud videos into point tubes for MAE-style reconstruction. Beyond the simplicity, we bring about more significant advantages in performance along with a new paradigm on parameter-efficient static-to-dynamic point cloud transfer learning.

\subsection{Parameter-Efficient Transfer Learning}
Motivated by the wide use of large pre-trained language models, efficient tuning has gained attention in Natural Language Processing (NLP). Recent studies extend this idea to vision. Existing methods are grouped under two umbrellas. The first group reuses and transfers knowledge from pre-trained source models. ORCA~\cite{shen2023cross} proposes a general workflow that aligns source and target embeddings and then fine-tunes the source model. MoNA~\cite{ma2024learning} formalizes the knowledge gap between modalities later. The second group learns a general model with additive modules or tokens. CoOp~\cite{zhou2022learning} applies prefix tuning to adapt CLIP~\cite{radford2021learning} to image recognition tasks. AIM~\cite{yangaim}  adapts an image model for video tasks. \revise{IDPT~\cite{zha2023instance} focuses on parameter-efficient transfer learning on 3D perception. DAPT~\cite{zhou2024dynamic} jointly refine prompts, cutting tunable parameters while boosting accuracy. Moving a step further, in this work, we address the more challenging task of adapting a pre-trained static 3D model without temporal knowledge to 4D perception.}

\section{METHOD}
\subsection{PointATA Workflow}
To clarify the notation, we first define several variables. A domain $\mathcal{D}$ is defined by a feature space $\mathcal{X}$ , a label space $\mathcal{Y}$, and a joint probability distribution $P(\mathcal{X}, \mathcal{Y})$. In the adaptation from static to dynamic point clouds, the static 3D domain $\mathcal{D}^s$ and the dynamic 4D domain $\mathcal{D}^d$ differ in both feature and label distributions. Thus, $\mathcal{X}^d \neq \mathcal{X}^s, \mathcal{Y}^d \neq \mathcal{Y}^s$, and $P^d\left(\mathcal{X}^d, \mathcal{Y}^d\right) \neq P^s\left(\mathcal{X}^s, \mathcal{Y}^s\right)$. Given dynamic point cloud samples $\left\{x_i^d, y_i^d\right\}_{i=1}^{n^d}$ drawn from $P^d$, our goal is to learn a model $m^d$ that maps each input $x^d$ to its correct label $y^d$. We achieve this by leveraging a pretrained transformer and adapters. Specifically, we assume access to a source model $m^s$ trained on static point cloud data $\left\{x_i^s, y_i^s\right\}_{i=1}^{n^s}$ from domain $\mathcal{D}^s$. Given a loss function $l$, we then build $m^d$ on top of $m^s$ to minimize $\mathbb{E}_{\left(x^d, y^d\right) \sim P^d}\left[l\left(m^d\left(x^d\right), y^d\right)\right]$. Model $m^d$ is composed of a frozen 3D model $m^s$ and learnable PVA with SCE. Consequently, our workflow in Fig.~\ref{pipeline} consists of two stages: (1) pre-training an embedder to align the feature distributions of static and dynamic point clouds as possible, and (2) 4D PETL to minimize the target loss.

\subsection{Stage1: Embedder Alignment}
Transferring knowledge across static and dynamic modalities is non-trivial because of two main challenges: dimensionality mismatch and metric unreliable. We argue that the embedder is the key to addressing both of them. Specifically, we can replace and reshape the embedder of dynamic point clouds to match the dimensionality of the pre-trained 3D model. Meanwhile, inspired by optimal transport dataset distances (OTDD)~\cite{alvarez2020geometric}, we compare feature–label pairs across domains, modelling labels as distributions over feature vectors. In this way, even when the label sets of two datasets are unrelated or disjoint (\textit{i.e.} ShapeNet~\cite{chang2015shapenet} and MSR-Action3D~\cite{5543273}), we can still compare the datasets across modalities and encourage the embedded 4D features to resemble the static 3D features, leading to more effective transfer learning.

Formally, let the 3D embedder be $f^s: \mathcal{X}^s \rightarrow \dot{\mathcal{X}}$, whose architecture follows mini-PointNet~\cite{qi2017pointnet}, while the 4D embedder be $f^d: \mathcal{X}^d \rightarrow \dot{\mathcal{X}}$ follows the P4DConv~\cite{fan21p4transformer}. In Stage 1, we train $f^d$ to minimize the distance between the joint distributions of 4D embeddings $\left(f^d\left(x^d\right), y^d\right)$ and 3D embeddings $\left(f^s\left(x^s\right), y^s\right)$. Considering that the labels are discrete, we represent each class label as a distribution over the corresponding features: $y \mapsto P(\dot{\mathcal{X}} \mid \mathcal{Y}=y)$. This maps both 3D and 4D label sets to the shared space of distributions on $\dot{\mathcal{X}}$. Then, we define the label distance $d_{\mathcal{Y}}\left(y^d, y^s\right)$ via the $p$-Wasserstein distance induced by the $l_2$ metric $\left\|\dot{x}^d-\dot{x}^s\right\|_2^2$ in $\dot{\mathcal{X}}$, enabling us to measure the distributional gap in $\dot{\mathcal{X}} \times \mathcal{Y}$:
\begin{equation}
d_{\dot{\mathcal{X}} \times \mathcal{Y}}\left((\dot{x}^d, y^d),\left(\dot{x}^s, y^s\right)\right)=\left(d_{\dot{\mathcal{X}}}(\dot{x}^d, \dot{x}^s)^p+d_{\mathcal{Y}}(y^d, y^s)^p\right)^{1/p} .
\end{equation}

\begin{algorithm}[tb]
  \caption{Class-Weighted Stochastic OTDD}
  \label{alg}
\begin{algorithmic}[1]
  \STATE {\bfseries Input:} 4D dynamic point cloud dataset $\{x^d, y^d\}$, number of 4D classes $K^d$, 3D static point cloud dataset $S=\{x^s, y^s\}$, subsample size $b$, subsample round $R$
  \FOR {each class $i \in [K^d]$ in the 4D dataset}
  \STATE Compute class weight $w_i = \frac{\text{number of 4D data in class } i}{\text{total number of 4D data}}$
  \STATE Generate data loader $D_i$ consisting of data in class $i$
  \ENDFOR
  \FOR{$i \in [K^d]$} 
  \FOR{$r \in [R]$}
  \STATE Subsample $b$ 4D data samples $D_{ir}$ uniformly at random from $D_i$
  \STATE Compute class-wise distance $d_{ir} = OTDD(D_{ir},S)$
  \ENDFOR
  \STATE Approximate class-wise OTDD by $d_i = \frac{1}{R}\sum_{i=1}^{R} d_{ir}$
  \ENDFOR
  \STATE Approximate OTDD by $d = \sum_{i=1}^{K^d} w_i \cdot d_i$
\end{algorithmic}
\end{algorithm}

\revise{Mathematically, this is formulated as minimizing the Wasserstein distance of order $p=2$ between the 4D and 3D joint distributions in the embedding space:}
\begin{equation}
\min _{f^d} W_2\left(P^d\left(\dot{\mathcal{X}}^d, \mathcal{Y}^d\right), P^s\left(\dot{\mathcal{X}}^s, \mathcal{Y}^s\right)\right).
\end{equation}

\revise{For orthogonal semantics, $P^d\left(y_i^d\right)$ and $P^s\left(y_j^s\right)$ have no overlap, so the optimization does not force $P^d\left(\dot{\mathcal{X}}^d \mid y_i^d\right)$ to map to any specific $P^s\left(\dot{\mathcal{X}}^s \mid y_j^s\right)$. Instead, it encourages $P^d\left(\dot{\mathcal{X}}^d \mid y_i^d\right)$ to inherit the structural properties of $P^s\left(\dot{\mathcal{X}}^s\right)$ while preserving its own semantic uniqueness.} Moreover, to reduce computational cost, we adopt the Sinkhorn~\cite{10.5555/2999792.2999868} and~\cite{shen2023cross}, lowering the complexity of OT from $\mathcal{O}\left(D^3 \log D\right)$ to $\mathcal{O}\left(D^2\right)$. As shown in Algorithm~\ref{alg}, this allows our 4D embedder invests a reasonable time budget but achieves significantly improved performance. We also illustrate the role of the Point Align Embedder in Section~\ref{abl}.

\subsection{Stage2: Efficient Adaptation}
To leverage large pre-trained 3D models for more challenging 4D video understanding, such as action recognition, we must bridge the intrinsic gap between static and dynamic point clouds. We regard the adapter architecture as a promising route toward it. Adapter~\cite{pmlr-v97-houlsby19a}, which were originally designed for parameter-efficient transfer learning in NLP. Indeed, an adapter module comprises a down-projection linear layer, a non-linear activation, and an up-projection linear layer. Adapters have achieved remarkable success in NLP because of three main advantages: (1) High parameter efficiency, as only a small set of parameters is task-specific; (2) Reaching on-par performance compared to full fine-tuning; (3) Mitigation of catastrophic forgetting. Our goal is to reproduce this success in 4D Vision. To this end, we introduce a new adapter tailored for spatio-temporal understanding, providing a capability that existing NLP-oriented adapters lack.

Typically, static point cloud models focus solely on spatial modeling~\cite{pang2022masked,chen2024pointgpt,yu2021pointbert}. The goal of our Point Video Adapter is to enable a pre-trained 3D model to grasp spatio-temporal cues in dynamic point clouds while adhering to a parameter-efficient principle. Compared with existing methods~\cite{11093957}, we focus on two practically crucial criteria: (1) \textbf{Smaller parameter footprint}: The parameter cost for every downstream task must remain small; Otherwise, the original purpose of adapters sink into oblivion and downstream tasks face a higher risk of overfitting. (2) \textbf{Easier implementation}: In practice, the model should be realizable with standard, highly-optimized deep-learning toolkits instead of introducing cumbersome or opaque modules. This requirement is essential for the advancement of robotic vision and embodied intelligence.

Guided by these considerations, we design the \texttt{Point Video Adapter} only using standard primitive operators. We add a spatio-temporal operator realized by a depth-wise separable convolution within the bottleneck. As the operator acts on the compressed low-dimensional features, \textit{e.g.} 128D, the PVA remains both parameter and compute efficient, formulated as:
\begin{equation}
\operatorname{\texttt{PVA}}(\mathbf{X})=\mathbf{X}+f\left(\operatorname{\texttt{DWConv}}\left(\mathbf{X} \mathbf{W}_{\text {down}}\right)\right) \mathbf{W}_{\text{up}},
\end{equation}
where \texttt{DWConv}$(\cdot)$ denotes the depth-wise separable convolution used for spatio-temporal modeling, $\mathbf{X}$ refers to input feature matrix, $\mathbf{W}_{\text {down}} \in \mathbb{R}^{d \times r}$ refers to the down projection layer, $\mathbf{W}_{up} \in \mathbb{R}^{r \times d}$ denotes the up-projection layer, and $f(\cdot)$ refers to the activation function. Notably, the design first learns local point features independently via grouped convolutions and then fuses cross-channel information with point-wise convolutions. Consequently, the model acquires layer-wise temporal understanding. This simultaneously enhances temporal modeling and reduces the computational cost of processing 4D data.

\begin{table*}[t]
\centering
\caption{Action recognition accuracy(\%)on the MSR-Action3D.}
\resizebox{\linewidth}{!}{
\begin{tabular}{lccccccc}
\toprule
Methods\qquad                 & Reference\qquad & 8 Frames\qquad\qquad & 12 Frames\qquad\qquad & 16 Frames\qquad\qquad & 24 Frames\qquad\qquad & 32 Frames\qquad\qquad & 36 Frames\qquad\qquad \\ \hline
\multicolumn{8}{c}{\textit{Supervised Learning}}                                   \\ \hline
MeteorNet~\cite{liu2019meteornet}\qquad       & ICCV2019\qquad  & 81.14\qquad\qquad    & 86.53\qquad\qquad     & 88.21\qquad\qquad     & 88.50\qquad\qquad &N/A\qquad\qquad & N/A\qquad\qquad   \\
PSTNet~\cite{fan2021pstnet}\qquad        & ICLR2021\qquad  & 83.50\qquad\qquad    & 87.88\qquad\qquad     & 89.90\qquad\qquad     & 91.20\qquad\qquad  &N/A\qquad\qquad& N/A\qquad\qquad  \\
P4Transformer~\cite{fan21p4transformer}\qquad   & CVPR2021\qquad  & 83.17\qquad\qquad    & 87.54\qquad\qquad     & 89.56\qquad\qquad     & 90.94\qquad\qquad &87.93\qquad\qquad&82.81\qquad\qquad    \\
Kinet~\cite{zhong2022no}\qquad         & CVPR2022\qquad  & 83.84\qquad\qquad    & 88.53\qquad\qquad     & 91.92\qquad\qquad     & 93.27\qquad\qquad&N/A\qquad\qquad& N/A\qquad\qquad    \\
PPTr~\cite{wen2022point}\qquad          & ECCV2022\qquad  & 84.02\qquad\qquad    & 89.89\qquad\qquad     & 90.31\qquad\qquad     & 92.33\qquad\qquad &N/A\qquad\qquad&N/A\qquad\qquad    \\
LeaF~\cite{10377208}\qquad          & ICCV2023\qquad  & 84.50\qquad\qquad    & N/A\qquad\qquad         & 91.50\qquad\qquad     & 93.84\qquad\qquad &N/A\qquad\qquad&N/A\qquad\qquad    \\
PST-Transformer~\cite{9740525}\qquad& TPAMI2023\qquad & 83.97\qquad\qquad    & 88.15\qquad\qquad     & 91.98\qquad\qquad     & 93.73\qquad\qquad &N/A\qquad\qquad&N/A\qquad\qquad    \\
X4D-SceneFormer~\cite{jing2024x4d}\qquad & AAAI2024\qquad & 86.47\qquad\qquad    & N/A\qquad\qquad         & 92.56\qquad\qquad     & 93.90\qquad\qquad &N/A\qquad\qquad&N/A\qquad\qquad    \\
MAMBA4D~\cite{liu2025mamba4d}\qquad                 & CVPR2025\qquad  & N/A\qquad\qquad        & N/A\qquad\qquad         & N/A\qquad\qquad         & 92.68\qquad\qquad &93.10\qquad\qquad& 93.23\qquad\qquad   \\ \hline
\multicolumn{8}{c}{\textit{Self-Supervised Learning (End-to-End Fine-Tuning)}}     \\ \hline
CPR~\cite{sheng2023contrastive}\qquad          & AAAI2023\qquad  & 86.53\qquad\qquad    & 91.00\qquad\qquad     & 92.15\qquad\qquad     & 93.03\qquad\qquad &N/A\qquad\qquad& N/A\qquad\qquad   \\
C2P~\cite{zhang2023complete}\qquad          & CVPR2023\qquad  & 87.16\qquad\qquad    & N/A\qquad\qquad         & 91.89\qquad\qquad     & 94.76\qquad\qquad   &N/A\qquad\qquad&N/A\qquad\qquad  \\
PointCMP~\cite{shen2023pointcmp}\qquad      & CVPR2023\qquad  & 89.56\qquad\qquad    & 91.58\qquad\qquad     & 92.26\qquad\qquad     & 93.27\qquad\qquad &N/A\qquad\qquad& N/A\qquad\qquad   \\
PointCPSC~\cite{sheng2023point}\qquad      & ICCV2023\qquad  & 88.89\qquad\qquad    & 90.24\qquad\qquad     & 92.26\qquad\qquad     & 92.68\qquad\qquad  &N/A\qquad\qquad&N/A\qquad\qquad   \\
MaST-Pre~\cite{shen2023masked}\qquad        & ICCV2023\qquad  & N/A\qquad\qquad        & N/A\qquad\qquad         & N/A\qquad\qquad         & 94.08\qquad\qquad &N/A\qquad\qquad& N/A\qquad\qquad   \\ \hline
\multicolumn{8}{c}{\textit{Adaptation for Pre-Trained Models}}                     \\ \hline
Point-CSA~\cite{11093957}\qquad           & CVPR2025\qquad  & 91.41\qquad\qquad    & 92.04\qquad\qquad     & 93.73\qquad\qquad     & 95.12\qquad\qquad &95.42\qquad\qquad& 95.42\qquad\qquad   \\
\rowcolor{gray!10}
Point-BERT+ATA\qquad          & -\qquad         & $\textbf{91.98}_{\color{blue}(+0.57)} $        & $\textbf{92.68}_{\color{blue}(+0.64)}$          &  $93.37_{\color{red}(-0.36)}$       & $95.28_{\color{blue}(+0.16)}  $   &$95.47_{\color{blue}(+0.05)}$&$96.16_{\color{blue}(+0.74)}$     \\
\rowcolor{gray!10}
Point-MAE+ATA\qquad           & -\qquad         &    $ 91.28_{\color{red}(-0.13)}$     &      $92.33_{\color{blue}(+0.29)} $    & $\textbf{94.42}_{\color{blue}(+0.69)}$          & $\textbf{95.62}_{\color{blue}(+0.50)}   $    &$\textbf{96.86}_{\color{blue}(+1.44)}$& $\textbf{97.21}_{\color{blue}(+1.79)}$  \\
\rowcolor{gray!10}
PointGPT-S+ATA\qquad & -\qquad         &  $ 91.28_{\color{red}(-0.13)}$ &$ \textbf{92.68}_{\color{blue}(+0.64)}$        & $94.07_{\color{blue}(+0.34)}$     &$ 95.28_{\color{blue}(+0.16)}$   &$95.12_{\color{red}(-0.30)}$&$95.12_{\color{red}(-0.30)}$  \\ \bottomrule
\end{tabular}}
\label{msr}
\end{table*}

\begin{table}[t]
\centering
\caption{3D Action Recognition Accuracy (\%) on NTU RGB+D 60 and NTU RGB+D 120. D. means depth. P. means Point. V. means voxel.}
\resizebox{\linewidth}{!}{
\begin{tabular}{lccccc}
\toprule
\multirow{2}{*}{Method}   & \multirow{2}{*}{Input} & \multicolumn{2}{c}{NTU 60} & \multicolumn{2}{c}{NTU 120} \\ \cline{3-6} 
                          &                        & Subject       & View       & Subject       & Setup       \\ \midrule
NTU Baseline~\cite{shahroudy2016ntu} & D.                  & N/A             & N/A          & 48.7          & 40.1        \\
PointNet++~\cite{qi2017pointnet++}       & P.                  & 80.1          & 85.1       & 72.1          & 79.4        \\
3DV~\cite{9157595}              & V.                  & 84.5          & 95.4       & 76.9          & 92.5        \\
3DV-PN++~\cite{9157595}   & V+P          & 88.8          & 96.3       & 82.4          & 93.5        \\
PSTNet~\cite{fan2021pstnet}           & P.                 & 90.5          & 96.5     & 87.0          & 93.8        \\
P4Trans.~\cite{fan21p4transformer}    & P.                  & 90.2          & 96.4       & 86.4          & 93.5        \\
PST-Trans.~\cite{9740525} & P.                  & 91.0          & 96.4       & 87.5          & 94.0    \\
\rowcolor{gray!10}
PointATA                  & P.                  & $\textbf{91.4}_{\color{blue}(+1.2)}$          & $\textbf{96.5}_{\color{blue}(+0.1)}$       & $\textbf{88.0}_{\color{blue}(+1.6)}$          & $\textbf{94.0}_{\color{blue}(+0.5)}$        \\ \bottomrule
\end{tabular}}
\label{ntuexp}
\end{table}

\subsection{Overall Interaction}
Till now, we have fused spatio-temporal cues with the 3D prior and completed the entire ``Align then Adapt” workflow. To provide our model with global spatial context, we introduce \texttt{Spatial Context Encoder (SCE)} implemented as an origin MLP. Spatial Context Encoder across layers share the same architecture but use distinct parameters. We formalize the overall interaction as:
\begin{equation}
\begin{aligned}
\mathcal{F}_{i+1}=\texttt{FFN}\left(\operatorname{Attention}\left(\mathcal{F}_i\right)\right)+\texttt{PVA}\left(\mathcal{F}_i\right)+\texttt{SCE}\left(\mathcal{F}_i\right),
\end{aligned}
\end{equation}
where $\mathcal{F}_i$ denotes 4D features in $i$-th layer, \texttt{FFN} denotes a feed-forward network whose weights, together with the attention parameters, are pre-trained on static point clouds and kept frozen during adaptation.

\section{EXPERIMENT}
Experiment settings are standardized in Section~\ref{expsett}. We evaluate the ``Align then Adapt” (PointATA) paradigm on 3D action recognition, 4D action segmentation, 4D semantic segmentation, \revise{gesture recognition and 4D scene flow prediction} from Section~\ref{action3d} to ~\ref{kitti}. Section~\ref{abl} analyzes architecture design, tuning strategies, adapters locations, training logs and feature visualization in detail. Following ~\cite{11093957}, we adopt PointBERT~\cite{yu2021pointbert}, Point-MAE~\cite{pang2022masked}, and PointGPT-S~\cite{chen2024pointgpt} as baselines to ensure fair comparison.

\subsection{Experiment Settings} 
\label{expsett}
We apply PointATA to three baselines: PointBERT~\cite{yu2021pointbert}, Point-MAE~\cite{pang2022masked}, and PointGPT-S~\cite{chen2024pointgpt}. Each has been pre-trained on ShapeNet~\cite{chang2015shapenet} for 300 epochs. Unless stated otherwise, all baselines share the same setup. The pre-trained weights stay frozen. In Stage 1, only the Point Align Embedder is updated for 10 epochs, during which the OTDD is minimized with a Wasserstein metric $p=2$, \revise{$b=32$}, Sinkhorn regularization $\varepsilon=0.1$. In Stage 2, the Point Align Embedder, the PVA\&SCE, and the task head are jointly tuned for 40 epochs. We use SGD with a learning rate that warms up to 0.01 in the first 10 epochs and decays by 0.1 at the 20th and 30th epochs. Dropout is set to 0.5. All the experiments are conducted on NVIDIA GeForce RTX 3090 GPUs.

\begin{table}[t]
\caption{Action segmentation on the HOI4D dataset. Acc means Accuracy (\%).}
\centering
\resizebox{\linewidth}{!}{
\begin{tabular}{lcccc}
\toprule
Method                                                        & Reference & Acc                                  & Edit                                 & F1@50                                \\ \hline
\multicolumn{5}{c}{\textit{Supervised Learning}}                                                                                                                                               \\ \hline
P4Transformer~\cite{fan21p4transformer} & CVPR2021  & 71.2                                 & 73.1                                 & 58.2                                 \\
PPTr~\cite{wen2022point}                & ECCV2022  & 77.4                                 & 80.1                                 & 69.5                                 \\
MAMBA4D~\cite{liu2025mamba4d}           & CVPR2025  & 85.5                                 & 91.3                                 & 85.5                                 \\ \hline
\multicolumn{5}{c}{\textit{Self-Supervised Learning (End-to-End Fine-Tuning)}}                                                                                                                 \\ \hline
P4Trans+C2P~\cite{zhang2023complete}    & CVPR2023  & 73.5                                 & 76.8                                 & 62.4                                 \\
PPTr+C2P~\cite{zhang2023complete}       & CVPR2023  & 81.1                                 & 84.0                                 & 74.1                                 \\
X4D~\cite{jing2024x4d}                  & AAAI2024  & 84.1                                 & 91.1                                 & 84.8                                 \\
CrossVideo~\cite{liu2024crossvideo}     & ICRA2024  & 83.7                                 & 86.0                                 & 76.0                                 \\ \hline
\multicolumn{5}{c}{\textit{Adaptation for Pre-Trained Models}}      \\ \hline
\rowcolor{gray!10} P4Trans+ATA               & -         & $\textbf{79.9}_{\color{blue}(+8.7)}$ & $\textbf{74.7}_{\color{blue}(+1.6)}$ & $\textbf{65.0}_{\color{blue}(+6.8)}$ \\ \bottomrule
\end{tabular}}
\label{hoi}
\end{table}

\subsection{3D Action Recognition: MSR-Action3D}
\label{action3d}
\textbf{Dataset.} We conduct experiments on MSR-Action3D~\cite{5543273}, which has 567 depth videos across 20 action classes. Each video averages 40 frames. Following prior work~\cite{fan21p4transformer,11093957}, we split the set into 270 training videos and 297 test videos. During fine-tuning, we use only point coordinates and each frame is randomly down-sampled to 2048 points. \revise{We divide the videos into clips of 8, 12, 16, 24, 32 or 36 frames and average the clip-level probabilities to obtain the final video-level accuracy.}

\textbf{Comparison Results.} We select an anchor frame every two frames and sample 64 anchor points from it. We set the temporal tube length to 3, the spatial radius to 0.3, and randomly sample 32 neighboring points. Table~\ref{msr} shows that our adaptation method surpasses both supervised and self-supervised baselines, highlighting the promise of transferring knowledge from static 3D pre-trained models to 4D perception. Our PointATA paradigm shows that naive fine-tuning may harm transfer, but aligning the feature distribution using Point Align Embedder can resolve this issue and benefit fundamentally. Recent studies in 2D Vision~\cite{shen2023cross,yangaim} reveal that heavy adapters distort pre-trained weights and yield sub-optimal results. By exploiting the lightweight but engineering-oriented PVA, we reduce this distortion and avoid falling into the same trap.

\begin{figure}[t]
\centering
% \hspace{-6mm}
\includegraphics[width=\columnwidth]{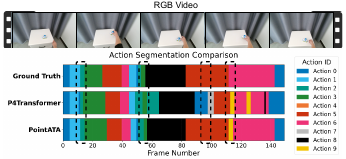}
\caption{\textbf{Visualization of action segmentation.} P4Transformer has a serious over-segmentation problem.}
\label{hoivis}
\end{figure}

\begin{table}[t]
\centering
\caption{4D semantic segmentation results on Synthia 4D dataset.}
\resizebox{\linewidth}{!}{
\begin{tabular}{lcccc}
\toprule
Method                  & Reference    & Input & Frame & mIoU (\%) \\ \hline
3D MinkNet14~\cite{choy20194d}   & CVPR2019  & voxel & 1     & 76.24     \\
4D MinkNet14~\cite{choy20194d}    & CVPR2019  & voxel & 3     & 77.46     \\ \hline
MeteorNet-M~\cite{liu2019meteornet}    & ICCV2019  & point & 2     & 81.47     \\
MeteorNet-L~\cite{liu2019meteornet}   & ICCV2019  & point & 3     & 81.80     \\
PSTNet~\cite{fan2021pstnet}         & ICLR2021  & point & 3     & 82.24     \\
P4Transformer~\cite{fan21p4transformer} & CVPR2021  & point & 1     & 82.41     \\
P4Transformer~\cite{fan21p4transformer}   & CVPR2021  & point & 3     & 83.16     \\
PST-Transformer~\cite{9740525} & TPAMI2022 & point & 1     & 82.92     \\
PST-Transformer~\cite{9740525} & TPAMI2022 & point & 3     & 83.95     \\
MAMBA4D~\cite{liu2025mamba4d}     & CVPR2025  & point & 3     & 83.35     \\
\rowcolor{gray!10}
PointATA                & -        & point & 1     & $\textbf{82.92}_{\color{blue}(+0.51)}$     \\
\rowcolor{gray!10}
PointATA                & -        & point & 3     & $\textbf{84.06}_{\color{blue}(+0.90)}$    \\ \bottomrule
\end{tabular}}
\label{syn4dexp}
\end{table}

\subsection{3D Action Recognition: NTU RGB+D}
\label{ntu}
\textbf{Dataset.} We conduct 3D action segmentation experiments on NTU RGB+D 60 and NTU RGB+D 120~\cite{shahroudy2016ntu}. The former comprises 56 k Kinect-v2 videos covering 60 action classes and 4 M frames recorded from 40 subjects with three cameras; evaluation follows cross-subject (20 vs 20 subjects) and cross-view (camera 1 for test, cameras 2–3 for training) protocols. The latter extends the collection to 114 k videos, 120 classes, 8 M frames, 106 subjects and 32 environmental setups, and adds a cross-setup split (16 setups for training, 16 for testing) alongside the conventional cross-subject partition. Consistent with prior research~\cite{fan21p4transformer}, we set the spatial radius of point 4D convolution to 0.1, sample clips of 24 frames, and train with a batch size of 32.

\begin{table}[t]
\caption{Gesture recognition accuracy (\%) on the SHREC'17.}
\centering
\resizebox{\linewidth}{!}{
\begin{tabular}{lcc}
\toprule
Methods      \qquad\qquad\qquad                       & Reference \qquad\qquad\quad            & Acc.(\%)            \\ \hline
\multicolumn{3}{c}{\textit{Supervised Learning}}                               \\ \hline
PLSTM-base~\cite{9157795}   \qquad\qquad \qquad            & CVPR2020  \qquad\qquad\quad            & 87.6                \\
PLSTM-early~\cite{9157795}   \qquad\qquad\qquad               & CVPR2020     \qquad\qquad\quad         & 93.5                \\
PLSTM-PSS~\cite{9157795}    \qquad\qquad\qquad               & CVPR2020  \qquad\qquad\quad            & 93.1                \\
PLSTM-middle~\cite{9157795}  \qquad\qquad\qquad               & CVPR2020    \qquad\qquad\quad          & 94.7                \\
PLSTM-late~\cite{9157795}   \qquad\qquad\qquad              & CVPR2020    \qquad\qquad\quad          & 93.5                \\
Kinet~\cite{zhong2022no}  \qquad\qquad \qquad                  & CVPR2022   \qquad\qquad\quad           & 95.2                \\ \hline
\multicolumn{3}{c}{\textit{Self-Supervised Learning (End-to-End Fine-Tuning)}} \\ \hline
PointCMP~\cite{shen2023pointcmp}     \qquad   \qquad  \qquad       & CVPR2023  \qquad\qquad\quad            & 93.3                \\
MaST-Pre~\cite{shen2023masked}\qquad\qquad  \qquad               & ICCV2023  \qquad\qquad\quad            & 92.4                \\ \hline
\multicolumn{3}{c}{\textit{Adaptation for Pre-Trained Models}}                 \\ \hline
Point-BERT+CSA~\cite{11093957}   \qquad \qquad  \qquad                & CVPR2025  \qquad\qquad\quad            & 96.2                \\
Point-MAE+CSA~\cite{11093957}  \qquad \qquad  \qquad                   & CVPR2025  \qquad\qquad\quad            & 95.2                \\
PointGPT-S+CSA~\cite{11093957} \qquad \qquad   \qquad                  & CVPR2025   \qquad\qquad\quad           & 96.5                \\
\rowcolor{gray!10}
Point-BERT+ATA   \qquad   \qquad \qquad                & -  \qquad\qquad\quad                   &  $\textbf{96.4}_{\color{blue}(+0.2)}$                   \\
\rowcolor{gray!10}
Point-MAE+ATA  \qquad  \qquad \qquad                  & -    \qquad\qquad\quad                 &  $\textbf{95.5}_{\color{blue}(+0.3)}$                  \\
\rowcolor{gray!10}
PointGPT-S+ATA    \qquad \qquad \qquad                 & - \qquad\qquad\quad                    &   $\textbf{96.5}_{\color{blue}(+0.0)}$                  \\ \bottomrule
\end{tabular}}
\label{sherc}
\end{table}

\begin{table}[t]
\caption{Comparison results on KITTI. Metrics are EPE, outlier ratio ( $>0.3\mathrm{~m}$ or $5\%$ ), Acc3DS and Acc3DR.}
\centering
\resizebox{\linewidth}{!}{
\begin{tabular}{lccccc}
\toprule
Methods  &Reference  & EPE3D & Acc3DS & Acc3DR & Outliers \\ \hline
FlowNet3D~\cite{liu2019flownet3d} &CVPR2019 & 0.169 & 0.254 & 0.579 & 0.789    \\
FLOT~\cite{puy2020flot}  & ECCV2020   & 0.110 & 0.419  & 0.721  & 0.486    \\
OGSFNet~\cite{ouyang2021occlusion} &CVPR2021   & 0.075 & 0.706  & 0.869  & 0.327    \\
FESTA~\cite{wang2021festa}    &CVPR2021  & 0.097 & 0.449  & 0.833  & N/A        \\
3DFlow~\cite{wang2022matters} &ECCV2022    & 0.073 & 0.819  & 0.890  & 0.261    \\
BPF~\cite{cheng2022bi}   &ECCV2022     & 0.065 & 0.769  & 0.906  & 0.264    \\
SCOOP~\cite{lang2023scoop}    &CVPR2023  & 0.063 & 0.797  & 0.910  & 0.244    \\
MSBRN~\cite{cheng2023multi}    &ICCV2023  & 0.044 & 0.044  & 0.950  & 0.208    \\
DiffFlow3D~\cite{liu2023difflow3d} &ArXiv2023 & 0.031 & 0.955  & 0.966  & 0.108    \\
\rowcolor{gray!10}
PointATA &-  & \textbf{0.021} & \textbf{0.978}  & \textbf{0.998}  & \textbf{0.088}    \\
$\Delta$    &-        & $\color{blue}-0.148$   & $\color{blue}+0.724$           & $\color{blue}+0.419$    & $\color{blue}-0.701$   \\ \bottomrule
\end{tabular}}
\label{flowest}
\end{table}


\textbf{Comparison Results.} Table~\ref{ntuexp} reports results on the NTU-RGBD dataset and shows that our method handles real-world point-cloud videos well, attaining 96.5\% accuracy. Despite its simplicity and light weight, it competes with elaborate extractors such as PSTNet~\cite{fan2021pstnet}, P4Transformer~\cite{fan21p4transformer} and PST-Transformer~\cite{9740525} by relying only on basic motion extraction and plain feature encoding, confirming that the proposed approach acquires useful cues for challenging video understanding.

\subsection{4D Action Segmentation: HOI4D}
\label{action4d}
\textbf{Dataset.}
We evaluate action segmentation on HOI4D~\cite{Liu_2022_CVPR}. The goal is to label every frame in point cloud videos. The official split has 2971 training and 892 test scenes. Each sequence contains 150 frames with 2048 points. We adopt the P4Transformer~\cite{fan21p4transformer} as the backbone. We report frame-level accuracy, segmental edit distance, and segmental F1 at IoU thresholds of 50\%.

\textbf{Comparison Results.} 
Table~\ref{hoi} makes great progress over the baseline~\cite{fan21p4transformer}. For P4Transformer, we replace its pre-training strategy with PointMAE~\cite{pang2022masked} and compare against several strong 4D pre-training methods. PointATA surpasses the original model and even some 4D-specific tuning schemes~\cite{zhang2023complete}, demonstrating accurate frame-level action classification. Our PointATA leverages the engineering-oriented PVA\&SCE to harvest temporal cues, guiding the model to attend to motion at the frame level. Moreover, once the complex and time-consuming data augmentations and pre-training costs of 4D and 3D backbones are taken into account, PointATA offers a favorable trade-off between accuracy and efficiency.

\textbf{Visualization Samples.} We visualize the ground-truth~\cite{Liu_2022_CVPR}, P4Transformer~\cite{fan21p4transformer} and our segmentations in Fig.~\ref{hoivis}. Consistent with~\cite{jing2024x4d}, P4Transformer suffers from severe over-segmentation, which we attribute to overfitting local appearance changes while neglecting motion continuity. In contrast, PointATA aggregates richer context, locates action boundaries more accurately and markedly reduces over-segmentation.

\begin{figure*}[t]
\centering
\includegraphics[width=.9\linewidth]{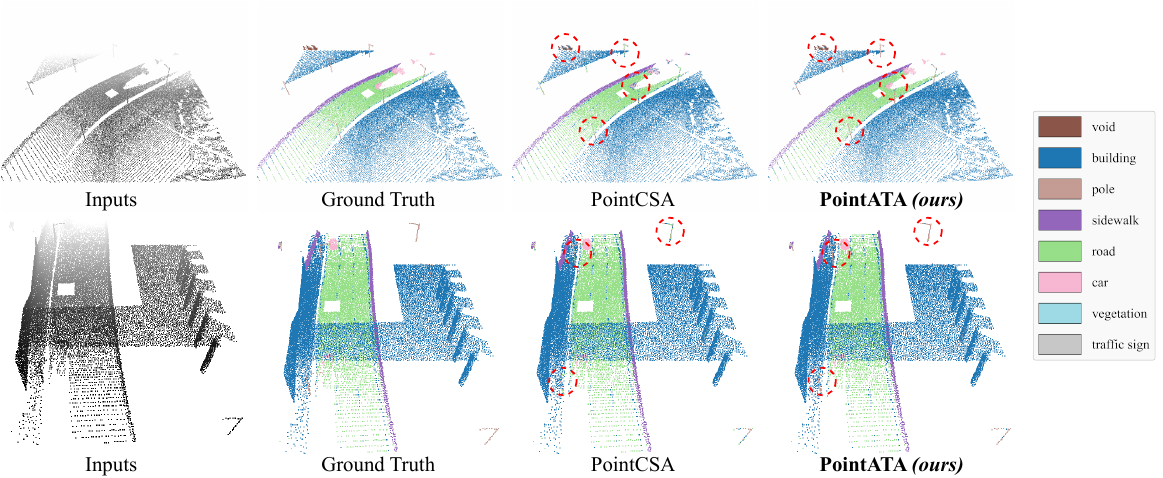}
\caption{\textbf{4D semantic segmentation visualization on the Synthia 4D dataset.} Key points are demarcated by red dashed circular bounding boxes.}
\label{syn4d}
\end{figure*}

\subsection{4D Semantic Segmentation: Synthia 4D}
\textbf{Dataset.} To verify that PointATA handles point-level tasks, we apply it to 4D semantic segmentation on Synthia 4D~\cite{choy20194d}. The dataset contains six driving sequences in which both objects and cameras move; each frame provides four stereo RGB-D images captured from a car roof. Following~\cite{liu2019meteornet}, we reconstruct point cloud videos and adopt the standard split: 19888 training, 815 validation and 1886 test frames. Consistent with~\cite{liu2019meteornet,choy20194d}, we sample clips of three consecutive frames. Note that, although single-frame segmentation is possible, exploiting temporal correlations improves scene understanding, segmentation accuracy and noise robustness. Performance is reported as mean Intersection over Union (mIoU).

\textbf{Comparison Results.} Table~\ref{syn4dexp} compares PointATA with CNN-based and Transformer-based methods. Without task-specific modules, PointATA consistently surpasses prior CNNs and Transformers on overall mIoU, yielding a 0.51 mIoU gain over our baseline~\cite{fan21p4transformer}. When the clip length is increased to three frames, the gap widens to 0.9 mIoU.

\textbf{Visualization Samples.} We present two segmentation results from the Synthia 4D dataset in Fig.~\ref{syn4d}. Our method can accurately segment more objects than the conventional PETL method.

\subsection{Gesture Recognition: SHREC'17}
\label{gesture}
\textbf{Dataset.} We utilize SHREC'17~\cite{desmedt:hal-01563505} for gesture recognition. It contains 2800 videos covering 28 gestures. Following~\cite{shen2023masked}, we split the dataset into 1960 training videos and 840 test videos. We keep the same model architecture as in MSR-Action3D~\cite{5543273}.

\textbf{Comparison Results.} As shown in Table~\ref{sherc}, our method surpasses both supervised and self-supervised baselines. With the ``Align then Adapt” strategy, PointGPT-S\cite{chen2024pointgpt}+ATA reaches 96.5\% accuracy, Point-BERT\cite{yu2021pointbert}+ATA achieves 96.4\%, and Point-MAE\cite{pang2022masked}+ATA attains 95.5\%. SHREC'17 is essentially a noisy, fine-grained task of counting fingers and locating joints. Success depends on structural hand cues, which is the exact strength of 3D pre-trained models. By adding the PVA, our method captures temporal dynamics and joint identities, complementing the frozen 3D backbone.

\subsection{4D Scene Flow Prediction: KITTI}
\label{kitti}
\revise{\textbf{Dataset.} To validate PointATA for 4D scene flow prediction, we evaluate it on KITTI~\cite{menze2015object}. The dataset has 200 training and 200 test samples, each with 4 frames of point cloud. We use the standard split and follow the evaluation protocol of FlowNet3D~\cite{liu2019flownet3d}. We report end-point error (EPE3D), strict 3D accuracy (Acc3DS), relaxed 3D accuracy (Acc3DR), and Outlier Ratio (Ourliers). An outlier is a point whose EPE exceeds 0.3m or 5\% of the ground-truth flow magnitude.}

\revise{\textbf{Comparison Results.} Experimental results in Table~\ref{flowest} demonstrate that PointATA attains statistically significant improvements across all evaluation metrics, surpassing 97\% accuracy on both Acc3DS and Acc3DR. The proposed point video adapter constitutes a critical component for scene flow prediction. By interleaving depth-wise convolutions with dedicated temporal-modeling blocks, it captures complex motion patterns with high fidelity. This architectural design systematically enhances the model’s robustness to rapid motion and severe occlusion.}

\begin{figure*}[t]
\centering
\includegraphics[width=1\linewidth]{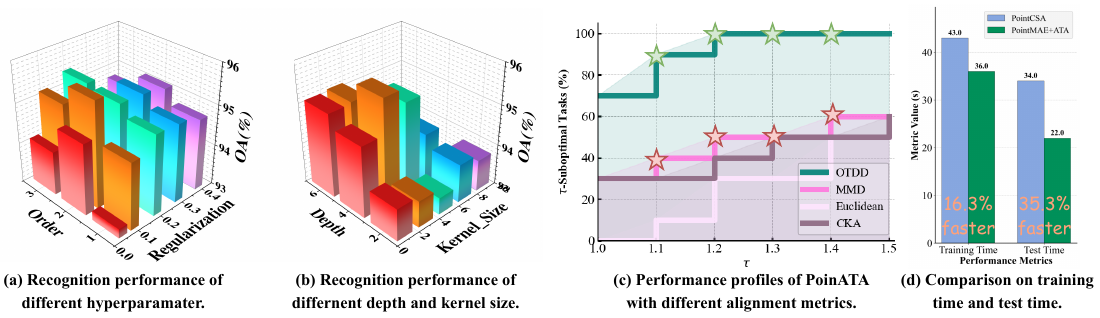}
\caption{\textbf{Ablation studies of PointATA under various experimental settings.}}
\label{r1abl4fig}
\end{figure*}

\subsection{Ablation Studies}
\label{abl}
\revise{In this subsection, we conduct extensive ablation studies to validate the effectiveness of the components and strategies proposed in PointATA. Unless otherwise specified, all ablation studies use PointMAE~\cite{pang2022masked} as the 3D backbone and are evaluated on the MSR-Action3D~\cite{5543273} dataset.} 

\revise{\textbf{Hyperparameter Sensitivity.} In Stage 1, we use the Class-Weighted Stochastic OTDD distance as the metric distance and treat the Wasserstein order $p$ and Sinkhorn regularization $\varepsilon$ as hyperparameters. Fig. \ref{r1abl4fig}(a) shows how the joint variation of these two parameters affects downstream task performance. Unlike regularization, continuously increasing the Wasserstein order does not necessarily improve downstream performance consistently. We argue that an overly complex Wasserstein distance may have negative effects and produce unstable gradients when label distribution differences are small.}

\begin{table}[t]
\caption{\revise{Training and inference advantages of PointATA over baseline. All values are recorded in seconds. Results are averaged over 10 runs. T. means time.}}
\centering
\resizebox{\linewidth}{!}{
\begin{tabular}{lcccc}
\toprule
Model        & Clip T.& Training T. & Test T. & GPU Hours \\ \hline
PointCSA     & 189.3     & 43                & 34            & 6.8       \\
\rowcolor{gray!10}
PointATA & \textbf{238.4}     & \textbf{36}                & \textbf{22}            & \textbf{5.6}       \\
\quad \textit{-Stage 1} &89.2&N/A&N/A&N/A\\
\quad \textit{-Stage 2} &149.2&N/A&N/A&N/A\\
$\Delta$            & $\color{blue}+25.9\%$   & $\color{blue}-16.3\%$           & $\color{blue}-35.3\%$    & $\color{blue}-17.6\%$   \\ \bottomrule
\end{tabular}}
\label{time}
\end{table}

\begin{table}[t]
\caption{Ablation studies on architecture designs. PAE denotes Point Align Embedder, and SCE denotes additional Spatial Context Encoders.}
\centering
\resizebox{\linewidth}{!}{
\begin{tabular}{lcccc}
\toprule
Model \qquad\qquad        & PAE\qquad\qquad           & SCE\qquad\qquad          & PVA\qquad\qquad          & Accuracy (\%)             \\ \hline
A0 \qquad\qquad      & $\times$\qquad\qquad     & $\times$\qquad\qquad     & $\times$\qquad\qquad     & $77.35_{\color{red}(baseline)}$                \\
A1  \qquad\qquad     & $\checkmark$\qquad\qquad & $\times$\qquad\qquad     & $\times$\qquad\qquad     & $84.51_{\color{blue}(+7.16)}$              \\
A2  \qquad\qquad     & $\checkmark$\qquad\qquad & $\checkmark$\qquad\qquad & $\times$\qquad\qquad     & $88.55_{\color{blue}(+11.2)} $               \\
A3  \qquad\qquad     & $\checkmark$\qquad\qquad & $\times$\qquad\qquad     & $\checkmark$\qquad\qquad & $95.12_{\color{blue}(+17.77)}$                \\
\rowcolor{gray!10}
A4 \textit{(ours)}\qquad\qquad & $\checkmark$\qquad\qquad & $\checkmark$\qquad\qquad & $\checkmark$\qquad\qquad & $\textbf{95.62}_{\color{blue}(+18.27)}$ \\ \bottomrule
\end{tabular}}
\label{arc}
\end{table}

\begin{table}[t]
\centering
\caption{\revise{Effects of Epoch Number and Sub-sampling strategy for Class-Weighted Stochastic OTDD on Transfer Learning Performance.}}
\resizebox{\linewidth}{!}{
\begin{tabular}{lcccc}
\toprule
Model & OTDD & Epoch Number &Sub-Sampling & Accuracy (\%) \\ \hline
B0    &  $\times$    &  N/A\qquad  &N/A   &   $92.68_{\color{red}(baseline)}$       \\
B1    &  $\checkmark$    & 2\qquad   &16  &   $93.37_{\color{blue}(+0.69)}$       \\
B2    &  $\checkmark$    & 4    &16 &   $94.42_{\color{blue}(+1.74)}$       \\
B3    &  $\checkmark$    & 6     &32&   $94.77_{\color{blue}(+2.09)}$       \\
B4    &  $\checkmark$    & 8   &32  &   $ 95.12_{\color{blue}(+2.44)}$      \\
\rowcolor{gray!10}
B5    &  $\checkmark$    & 10   &32 &   $95.62_{\color{blue}(+2.94)}$       \\
B6    &  $\checkmark$    & 12   &64 &   $95.62_{\color{red}(tradeoff)}$       \\
\bottomrule
\end{tabular}}
\label{otddabl}
\end{table}

\begin{table}[t]
\caption{Scaling behavior of 3D backbones with PointATA. F. means frames per clip.}
\centering
\resizebox{\linewidth}{!}{
\begin{tabular}{lcccc}
\toprule
Model                      & Reference\qquad                   & Points\qquad & Clips\qquad & Accuracy (\%) \\ \hline
\multirow{4}{*}{PTM~\cite{10604912}}       & \multirow{4}{*}{TCSVT2024}\qquad  & 2048\qquad   & 32 F.\qquad & 95.12    \\
                           &                             & 4096\qquad   & 32 F.\qquad & 95.47    \\
                           &                             & 2048\qquad  & 36 F.\qquad & 96.51    \\
                           &                             & 4096\qquad  & 36 F.\qquad & 96.86    \\ \hline
\multirow{4}{*}{PointM2AE~\cite{zhang2022point}} & \multirow{4}{*}{NeurIPS2022}\qquad & 2048\qquad   & 32 F.\qquad & 94.42    \\
                           &                             & 4096\qquad   & 32 F.\qquad & 94.77    \\
                           &                             & 2048\qquad  & 36 F.\qquad & 95.47    \\
                           &                             & 4096\qquad  & 36 F.\qquad & 96.16    \\ \bottomrule
\end{tabular}}
\label{scaling}
\end{table}

\revise{\textbf{Metric Distance.} To demonstrate the superiority of our proposed Algorithm \ref{alg}, we compare its performance profile~\cite{dolan2002benchmarking} against Maximum Mean Discrepancy (MMD)~\cite{gretton2006kernel}, Centered Kernel Alignment (CKA)~\cite{kornblith2019similarity}, and Euclidean distances on downstream tasks, following the protocol of ORCA~\cite{shen2023cross}. Fig. ~\ref{r1abl4fig}(c) demonstrates that our transfer metric distance exhibits superior generalization, achieving better performance across more downstream tasks. Euclidean distance performs worst in transfer learning because it is a degenerate case of Wasserstein distance. When the empirical distribution is not uniform, the sampled points have unequal weights. Euclidean distance cannot prioritize samples under this condition, whereas our method can.}

\revise{\textbf{Computational Advantages.} Fig.~\ref{r1abl4fig}(d) and Table~\ref{time} show that PointATA outperforms the baseline in both training and inference. Under the same setting, PointATA improves inference speed by 35.3\% and reduces GPU burden by 17.6\% on average. Indeed, PointATA is a two-stage transfer method and most computation is spent on cross-modal alignment in Stage 1. Considering this, our SCE and PVA modules run faster and use less memory than the standard adapters.}

\revise{\textbf{Depth \& Kernel Size.} Fig.~\ref{r1abl4fig}(b) quantifies the sensitivity of transfer learning performance to kernel size and PVA depth. When the kernel size is too small, the receptive field becomes insufficient to encode point level motion, and downstream accuracy degrades by 1.3–2\%. Monotonically increasing the adapter depth initially yields consistent gains. However, beyond four blocks the trend reverses, indicating an optimal capacity threshold. We hypothesise that a deeper PVA asymptotically approaches the representational power of the full baseline, thereby re-introducing the overfitting bias that PointATA is explicitly designed to suppress. These observations corroborate our core motivation: a parsimonious yet expressive adapter outperforms naive parameter scaling.
}

\textbf{Architecture Design.} We propose PointATA to equip a static point-cloud pre-training model with spatio-temporal cues from three components: Point Align Embedder, Spatial Context Encoder, and Point Video Adapter. We conduct ablation studies on these three components in Table~\ref{arc}. Without any of these components, the model has only a very limited capacity to perceive dynamic point clouds (77.35\%). After adding PA and PVA, the model’s performance improves markedly (+17.77\%). Introducing SCE unlocks the model’s full potential, enabling it to surpass not only comparable competitors on MSR-Action3D but also several pre-training models designed specifically for 4D perception.

\begin{figure}[t]
\centering
% \hspace{-6mm}
\includegraphics[width=1\columnwidth]{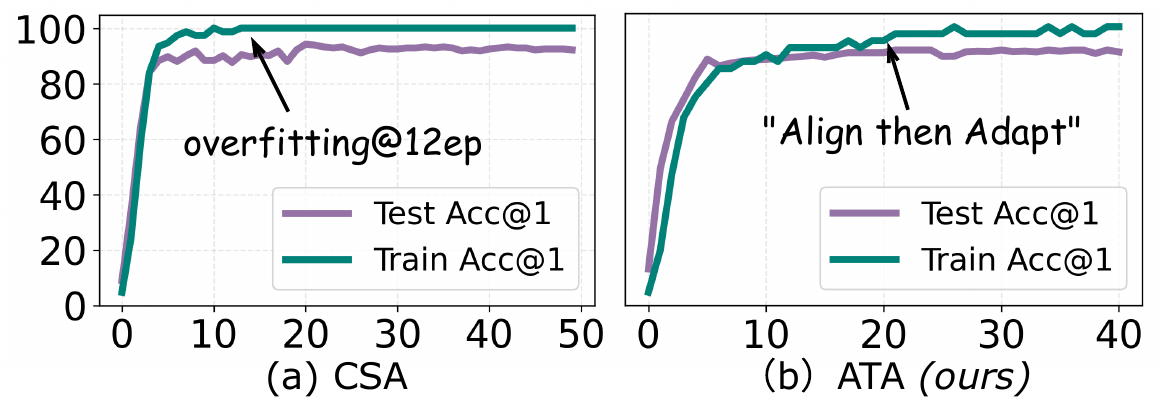}
\caption{\textbf{Training log visualization.} We consider 50 epochs for tuning is excessive. For a fair comparison with naive tuning, we shorten the schedule to 10 epochs for the embedding layer followed by 40 epochs of parameter-efficient fine-tuning.}
\label{log}
\end{figure}

\begin{figure}[t]
\centering
\includegraphics[width=0.95\columnwidth]{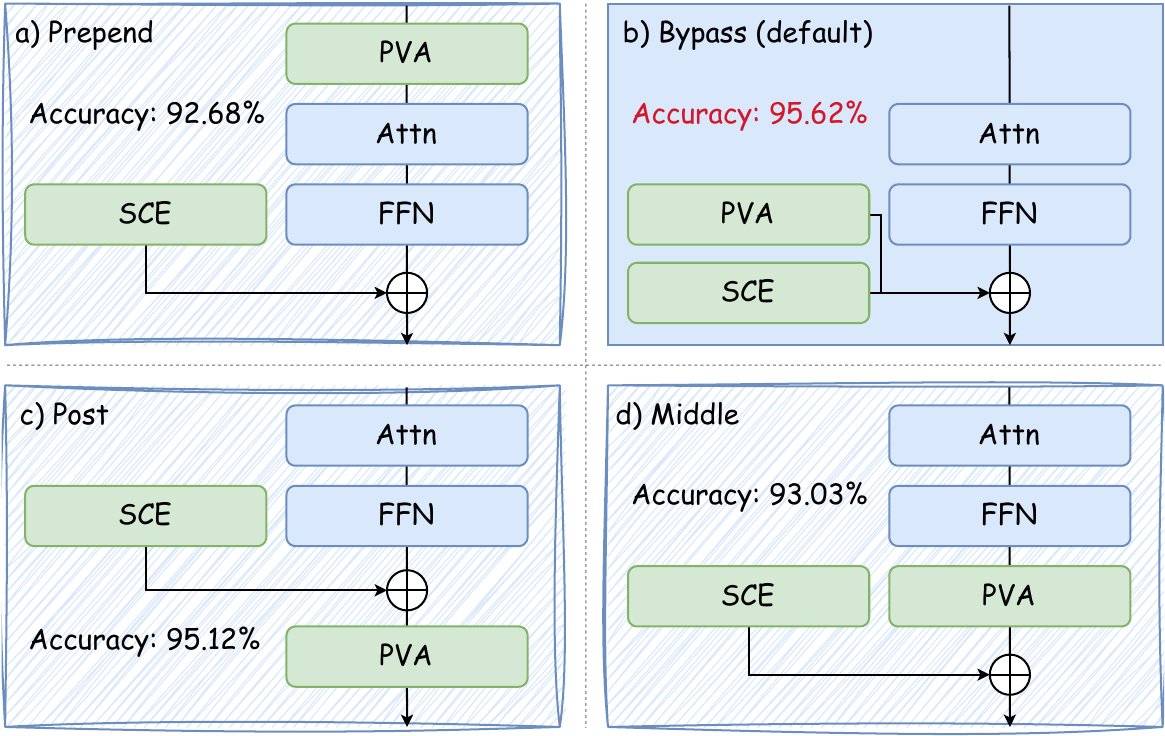}
\caption{\textbf{Different location of Point Video Adapter and Spatial Context Encoder.}}
\label{loc}
\end{figure}

\revise{\textbf{Proportion of Stage 1.} To rigorously verify that the cross modal alignment strategy in Stage 1 guarantees a lower target-task error, we conduct a controlled ablation. As reported in Table~\ref{otddabl}, adopting OTDD as the distance metric reproducibly elevates downstream performance. However, the marginal gain saturates after approximately 10 epochs. On the other hand, the default $b=32$ achieves a balanced trade-off. It maintains alignment stability comparable to $b=32/64$ while avoiding the excessive computational cost of larger $b$. Since PETL explicitly trades off accuracy against efficiency, limiting Stage 1 to 20\% of the total transfer learning budget is not only justified but optimal.}

\begin{figure}[t]
\centering
\includegraphics[width=0.8\columnwidth]{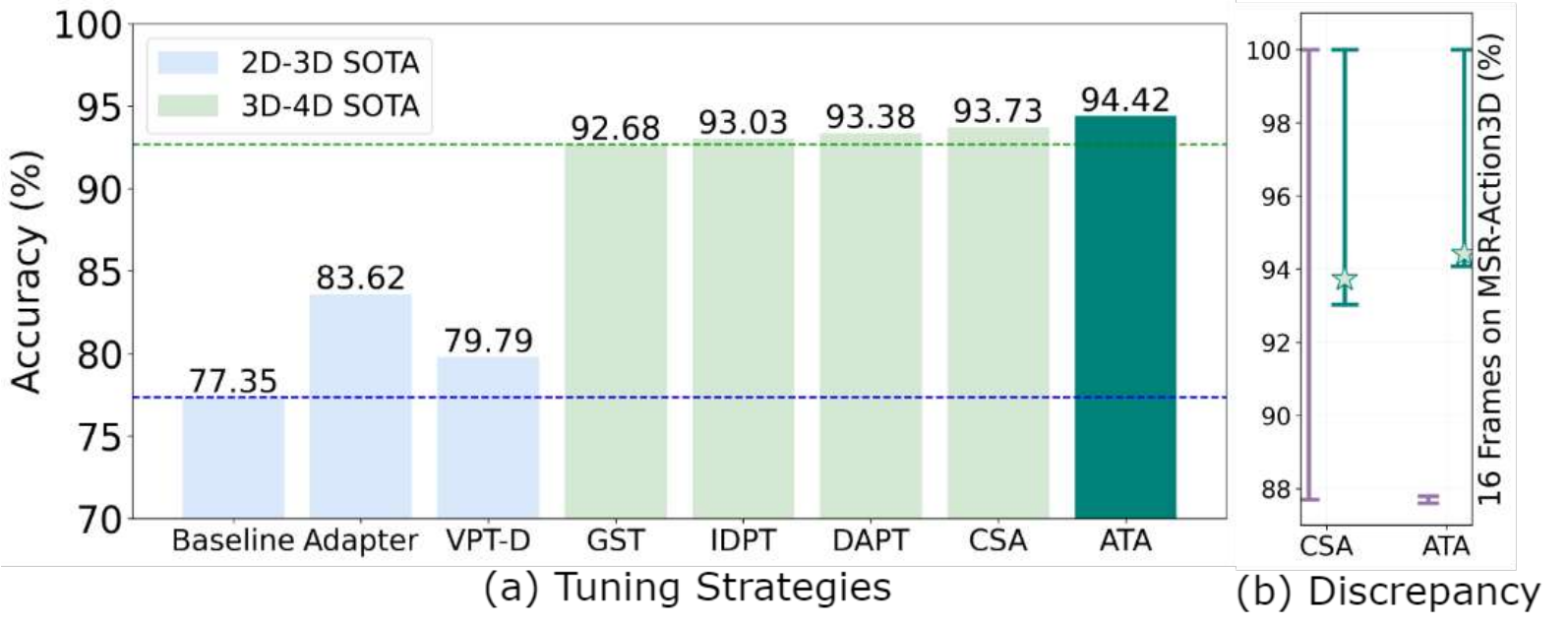}
\caption{\textbf{(a) Comparison of different tuning strategies. (b) Discrepancy between training and test accuracy under 16 frames on MSR-Action3D.}}
\label{tuning}
\end{figure}

\begin{figure}[t]
\centering
\includegraphics[width=.9\columnwidth]{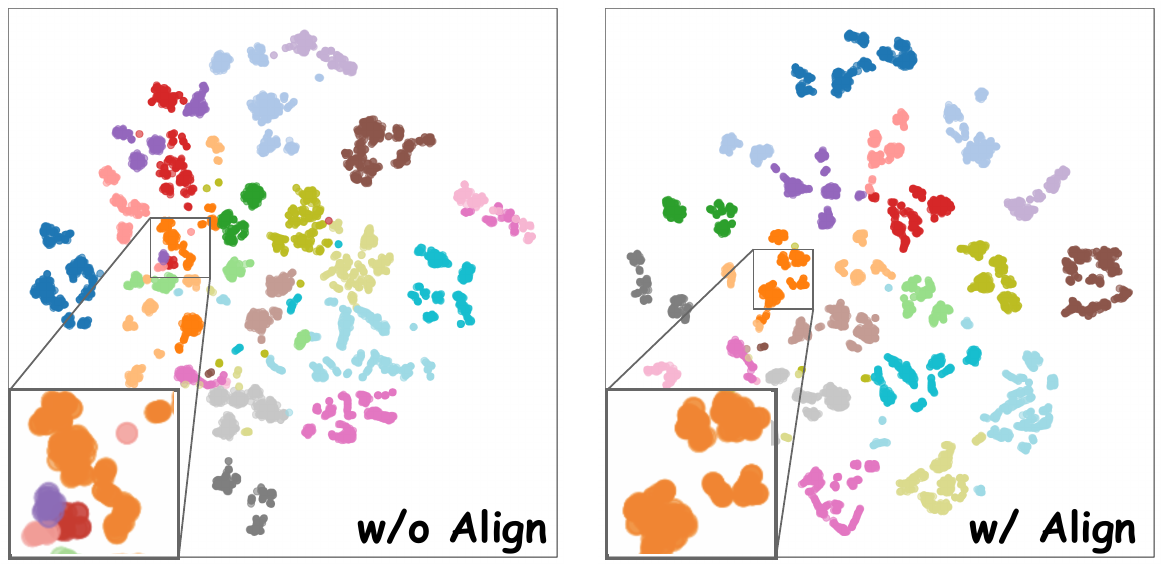}
\caption{\textbf{t-SNE visualization showing that aligning static and dynamic clouds with the Point Align Embedder is essential.} Smaller indices indicate tighter clusters. After alignment, intra-class distances shrink and inter-class gaps widen.}
\label{tsne}
\end{figure}

\revise{\textbf{Scaling and Generalization Properties.} We evaluate PointATA on additional 3D backbones to assess its scaling and generalization. Following community feedback, we select PTM~\cite{10604912} and PointM2AE~\cite{zhang2022point} as the foundation models. Table~\ref{scaling} shows that longer clips and denser points improve recognition when ATA is enabled. These results confirm that PointATA scales beyond small backbones and short clips, meeting the requirements of a general transfer learning paradigm.
}

\textbf{Training Log Visualization.} To confirm our motivation, we plot the training logs in Fig.~\ref{log}. Under naive adapter tuning, the model reaches 100\% training accuracy at Epoch 12, yet test accuracy improves slowly afterwards. The bar in Fig.~\ref{tuning}(b) represents the gaps between training and test accuracy.~\cite{11093957} tuning reaches 100\% training accuracy but only 87.8\% test accuracy (12.2\% train-test gap). For a model training over 50 epochs, such early overfitting is unacceptable.  Fortunately, with our ``Align then Adapt” paradigm, the overall training process is more stable and test accuracy rises smoothly. PointATA achieves 87.9\% training accuracy and 87.7\% test accuracy (merely 0.2\% gap) at the same epoch. Even after only 40 epochs, our method outperforms naive tuning counterparts. 

\textbf{Adapters Location.} We aim to identify the optimal insertion location for PVA and SCE so that the model can better integrate spatio-temporal features. Specifically, we conduct experiments at four locations: (a) Bypass, (b) Prepend, (c) Post, and (d) Middle, as shown in Fig.~\ref{loc}. PointATA performs worse when placed at prepend or post, suggesting that a serial adapter hampers the pre-trained model’s ability to process static information. Results at the bypass location indicate that a parallel design is preferable. This choice aligns with intuition: Static and dynamic cues are best handled by separate branches whose features are later fused, allowing the model to benefit from both sources.

\textbf{Tuning Strategies.} To demonstrate the superiority of our ``Align then Adapt” paradigm, we evaluate several tuning methods while keeping the pre-trained backbone frozen. As shown in Fig.~\ref{tuning}(a), \revise{ATA significantly outperforms the vanilla adapter~\cite{pmlr-v97-houlsby19a}, VPT-Deep~\cite{jia2022visual}, GST~\cite{liang2025parameter}, IDPT~\cite{zha2023instance}, DAPT~\cite{zhou2024dynamic} and CSA~\cite{11093957}.} This indicates that ATA enables effective cross-modal transfer for point cloud perception. PVA and SCE successfully captures temporal information while reusing the frozen backbone and keeping the overall model lightweight and replicable.

\textbf{t-SNE Visualization.} To qualitatively assess how Point Align Embedder helps the model grasp dynamic point cloud sequences, we visualize the learned features with t-SNE. As shown in Fig.~\ref{tsne}, after static-dynamic alignment, the features form tighter and more distinct clusters than naive adapter tuning. This confirms our second motivation: 4D PETL without any pre-alignment is inadequate.

\section{CONCLUSION}
In this paper, we identify and analyze two key limitations in transferring 3D knowledge to 4D perception, then proposing corresponding solutions. We present PointATA, the first ``Align then Adapt” paradigm for static-to-dynamic point cloud adaptation. The pipeline splits cross-modal transfer into two stages: In Stage 1, we employ OTDD to measure the distribution gap between 3D and 4D datasets and train the 4D embedder; In Stage 2, we integrate the Point Video Adapter and Spatial Context Encoder to capture both geometric and dynamics in 4D videos. \revise{Our method outperforms existing methods on 3D action recognition, 4D action recognition, gesture recognition and 4D scene flow prediction}, demonstrating the effectiveness of the PointATA paradigm. \revise{Owing to our limited resources, we acknowledge that PointATA may exhibit insufficient modeling capacity in certain outdoor scenarios, and we will address this issue in our future work. By validating PointATA on multiple visual benchmarks, we hope our work could inspire the community to rethink fine-tuning mechanism in 4D Vision.}

\bibliographystyle{IEEEtran}
\bibliography{tmm.bib}

\vfill

\end{document}